\documentclass{article} 
\usepackage{abc2027_conference,times}

\usepackage{comment}
\usepackage[utf8]{inputenc}
\usepackage[T1]{fontenc}
\usepackage[expansion=false]{microtype}

\usepackage{amsmath}
\usepackage{amssymb}
\usepackage{amsthm}
\usepackage{mathtools}
\usepackage{mathtools}
\usepackage{arydshln}
\usepackage{bm}
\usepackage{nicefrac}

\usepackage{graphicx}
\usepackage{booktabs}
\usepackage{float}
\usepackage{xspace}
\usepackage[dvipsnames]{xcolor}
\usepackage{booktabs}  
\usepackage{makecell}
\usepackage{adjustbox}

\usepackage{hyperref}
\usepackage{url}

\usepackage{algorithm}
\usepackage{algpseudocode}

\theoremstyle{plain}
\newtheorem{theorem}{Theorem}[section]
\newtheorem{lemma}[theorem]{Lemma}
\newtheorem{corollary}[theorem]{Corollary}
\newtheorem{proposition}[theorem]{Proposition}

\theoremstyle{definition}

\newtheorem{remark}[theorem]{Remark}

\newcommand{\polar}{\mathrm{polar}}

\newcommand{\softmax}{\mathrm{softmax}}

    \newcommand{\MuonIO}{\textrm{MuonIO}}

\input{Definitions}

\abcfinalcopy
\title{\MuonIO: Principled Norm-Aware Descent for\\Embedding Tables
       and Language Model Heads}

\author{%
  Linkai Ma\thanks{Equal contribution.}
  \hspace{0.4em}
  Xinyu Luo\footnotemark[1]
  \hspace{0.4em}
  Mengbo Wang
  \hspace{0.4em}
  Ananth Grama
  \hspace{0.4em}
  Petros Drineas
  \hspace{0.4em}
  Brian Bullins
  \\
  Department of Computer Science, Purdue University
  \\
  \texttt{\{ma856, luo466, wang4887, ayg, pdrineas, bbullins\}@purdue.edu}
}

\begin{document}

\maketitle
\lhead{}
\renewcommand{\headrulewidth}{0pt}
\begin{abstract}
The Muon optimizer derives its update rule for hidden linear layers by solving
 a local linearization of the loss penalized by the spectral norm,
motivated by an RMS-stability argument for dense linear layers.
Standard Muon implementations, however, exclude the input (embedding table) and output (language model head) layers from this principled
treatment, for which they use AdamW instead.
We present $\MuonIO$, a single Muon-style update
for both of these layers.
For the language model head $\Lb \in \RR{V}{d}$, we motivate the use of the $2\to\infty$ operator norm, due to the Lipschitz continuity of the softmax output geometry, while
for the embedding table $\Eb \in \RR{d}{V}$, we draw on the $1 \to 2$ operator norm, based on the one-hot input geometry identified by~\citet{bernstein2025modular}. 
The identity
$\|\Lb\|_{2\to\infty}=\|\Lb^\top\|_{1\to2}$ then puts both matrices in the
same vocabulary-oriented geometry: $\MuonIO$ applies a single normalized-vector rule,
which appears as column normalization for $\Eb$ and row normalization for
$\Lb$. 
Empirical evaluations demonstrate the effectiveness of our approach, with MuonIO reducing I/O optimizer state memory by 50\% and I/O update FLOPs by $\sim$46\% compared to Muon for 1B LLaMA pretraining on C4, while also improving validation perplexity.
\end{abstract}
\section{Introduction}

Gradient-based optimization of neural networks has long been dominated by
Adam~\citep{kingma2014adam} and its descendants. The Muon
optimizer~\citep{jordan2024muon} revisited a foundational question: what is
the right geometry for determining the direction of a weight update? Building
on ideas from modular duality~\citep{bernstein2024old,bernstein2025modular},
Muon chooses its update by minimizing a local first-order approximation of
the loss while penalizing the update in the \emph{spectral norm geometry}. This geometry
is appropriate for hidden linear layers~\citep{yang2023spectral} and yields the
polar factor update, which is
computed in practice by a Newton--Schulz iteration rather than by an explicit
SVD~\citep{jordan2024muon,bernstein2025modular}. Muon's practical impact has
been rapid: it achieved a $1.35\times$ speedup over the previous NanoGPT
training record~\citep{jordan2024muon}, has been used in large-scale models
such as Moonlight~\citep{liu2025muon} and Kimi K2~\citep{team2025kimi}, and
has inspired variants such as HTMuon~\citep{pang2026htmuon}, NorMuon~\citep{li2025normuon}, Polar Express~\citep{amsel2026the}, AdaMuon~\citep{si2025adamuon}, Scion~\citep{pethick2025training}, Dion~\citep{ahn2025dion}, and Dion2~\citep{ahn2025dion2}.

Despite this progress, standard Muon implementations still exclude two
important layer types from the spectral norm update: the input
\textbf{embedding layer} (\texttt{embed}) and the output
\textbf{language model head} (\texttt{lm\_head}). Instead, both are typically routed
to AdamW, and as~\citet{jordan2024muon} explicitly note, 
this exception is not fully explained by the existing theory: \vspace{-0.0cm}
\begin{quote}
\emph{``That such dynamics are also different for the output layer does
not seem to follow from the theory, and is instead driven by empirics.''}
\end{quote}
\vspace{-0.0cm}
Similarly, Bernstein's derivation note~\citep{bernstein2025deriving}
observes that the dense layer stability argument ``\emph{separates the Linear
layer from, say, an Embedding layer that takes one-hot inputs.}'' While prior work has identified the norm induced by one-hot embedding lookups~\citep{bernstein2024old,bernstein2025modular}, and column/row normalization updates have been proposed on other grounds~\citep{pethick2025training,glentis2026scale,lau2026symmetry}, there has yet to be established a principled norm-aware approach that handles both vocabulary-facing layers with a single update.

\subsection{Our Contribution}
In this work, we take the embedding layer geometry
\citep{bernstein2024old,bernstein2025modular} as a starting point and present
\textbf{MuonIO}, a single Muon-style update for the two vocabulary-facing
matrices. Prior work identifies the $1\to\mathrm{RMS}$ norm and its
column-normalized direction for \texttt{embed},
which, under our $\ell_2$ convention, is the $1\to2$ norm up to a fixed dimension factor. For
\texttt{lm\_head}, we instead study the change in the probability distribution
after softmax. A tight total-variation stability bound identifies the
$2\to\infty$ norm of the weight perturbation. These operator norms are related
by transposition:
$\|\Lb\|_{2\to\infty}=\|\Lb^\top\|_{1\to2}$. Once both parameters are stored in vocabulary-major form,
Corollary~\ref{cor:io-normalized-step} gives the same normalized-vector update
for both. In weight-tied models where
$\Lb=\Eb^\top$~\citep{press2017using,inan2016tying}, it acts directly on the
single shared parameter.

Optimizing \texttt{embed} and \texttt{lm\_head} with AdamW also carries concrete computation and memory cost. These two layers are among the largest parameter matrices in a language model: each has $Vd$ entries, with $V$ being the vocabulary size and $d$ the hidden dimension. AdamW maintains two optimizer state moments per parameter, whereas $\MuonIO$ keeps only a single momentum buffer. In terms of computation, AdamW takes $13Vd$ FLOPS per matrix while $\MuonIO$ takes $7Vd+3V$. Routing these two layers to AdamW therefore roughly doubles the optimizer state memory and arithmetic cost on the model's biggest tensors. Further details are included in Appendix~\ref{app:complexity}.

We evaluate MuonIO by pretraining LLaMA-family models at the 60M, 130M and 1B scales. Replacing the
AdamW treatment of the two I/O layers with one geometry-matched update halves their persistent optimizer state and, at the same time, yields improved validation perplexity over a strong Muon
baseline that already tunes separate learning rates for the input, output, hidden, and gain/bias parameters.

\subsection{Related Work}\label{sec:related}

The work on modular
duality~\citep{bernstein2025modular} gives a column normalization update
for \texttt{embed}, but does not provide a corresponding treatment of
\texttt{lm\_head}. Other optimizer frameworks have nevertheless proposed
row-normalized output layer updates, and recent work has applied
vocabulary-wise normalization specifically to language model heads, motivated
by linear-minimization oracles, empirical gradient variance, and equivariance
\citep{pethick2025training,glentis2026scale,lau2026symmetry}, while
Ember~\citep{shah2026token} treats both layers specially using factorized second moments. Another related
line of work observes, both empirically and theoretically, that \texttt{embed}
benefits from a learning rate distinct from the hidden
layers~\citep{hayou2025optimal,he2026one,kalra2026quantifying}. Additionally,~\citet{nair2025softmax} establishes tight Lipschitz bounds for the softmax function under matched $\ell_p$ norms. For a more detailed discussion, see Appendix~\ref{app:related-work}.

\section{Background and Notation}\label{sec:background}

\subsection{Notation}

Scalars are denoted by lowercase letters (e.g., $x$); vectors by bold lowercase letters (e.g., $\xb$); matrices by bold uppercase letters (e.g., $\Wb$). For a matrix $\Ab \in \RR{m}{n}$ we write $\|\Ab\|_F$ for the Frobenius norm, $\twonorm{\Ab}$ for the spectral norm, $\|\Ab\|_*$ for the nuclear norm, and $\langle \Ab, \Bb\rangle = \Tr{\Ab^\top\Bb}$ for the Frobenius inner product. We write $\Ab_{:,t}$ for the $t$-th column of $\Ab$ and $\Ab_{t,:}$ for its $t$-th row. We write $\Ib_k$ for the $k\times k$ identity matrix (dropping the subscript when the dimension is clear), and $[V]$ for $\{1, 2, \ldots, V\}$. We write $\Delta^V := \left\{p \in \R^V_{\ge 0}: \sum_{t=1}^V p_t = 1\right\}$ for the probability simplex over $V$ outcomes.

\subsection{Steepest Descent Under a Norm}\label{subsec:steepest-descent}

Let $\mathcal{L}(\Wb)$ be a loss function and let $\Gb_k = \nabla
\mathcal{L}(\Wb_k)$ denote the gradient at iterate $\Wb_k$. For a given
norm $\|\cdot\|$ on the weight matrices, the
\textbf{steepest descent step} of size $\lambda > 0$ minimizes a penalized
local linearization. The following generic result is
\citet[Proposition~1]{bernstein2024old}, stated for matrix spaces under the
Frobenius inner product. We use the reciprocal of their sharpness parameter so that
$\lambda$ is a step size.

\begin{proposition}[Penalized steepest descent]
\label{prop:anthology-steepest-descent}
Let $\Gb\in\RR{m}{n}$ and let $\|\cdot\|$ be a norm on $\RR{m}{n}$, with dual
norm $\|\Gb\|^\dagger := \max_{\|\Tb\|\le 1}\langle\Gb,\Tb\rangle$.
For $\lambda>0$,
\begin{equation}\label{eq:steepest-descent}
  \operatorname*{argmin}_{\Delta\Wb}
  \left\{
    \langle\Gb,\Delta\Wb\rangle
    +\frac{1}{2\lambda}\|\Delta\Wb\|^2
  \right\}
  =
  \left\{
    -\lambda\|\Gb\|^\dagger\Tb:
    \Tb\in\operatorname*{argmax}_{\|\Tb\|=1}
    \langle\Gb,\Tb\rangle
  \right\}.
\end{equation}
\end{proposition}

\textbf{Gradient descent (Frobenius geometry).}
Applying Proposition~\ref{prop:anthology-steepest-descent} with
$\|\cdot\| = \|\cdot\|_F$ yields $\Delta\Wb^\star = -\lambda \Gb$, the
ordinary gradient descent step.

\textbf{Muon (spectral norm geometry).}
Taking $\|\cdot\| = \twonorm{\cdot}$ and letting
$\Gb = \Ub\bm{\Sigma}\Vb^\top$ be the thin SVD of the gradient,
Proposition~\ref{prop:anthology-steepest-descent} gives the following choice of solution:
\begin{equation}\label{eq:muon-step}
  \Delta\Wb^\star = -\lambda\, \|\Gb\|_* \polar(\Gb), \qquad \polar(\Gb) := \Ub\Vb^\top.
\end{equation}

\subsection{Operator Norms}\label{subsec:op-norms}

For $p, q \in [1,\infty]$ and $\Ab \in \RR{m}{n}$, the
\textbf{$p\to q$ operator norm} is $ \|\Ab\|_{p\to q} := \max_{\|\xb\|_p = 1}\|\Ab\xb\|_q.$
For example, the spectral norm satisfies $\twonorm{\Ab} = \|\Ab\|_{2\to 2}$. \citet[Proposition~8]{bernstein2024old} establish the following induced-norm identities.

\begin{proposition}
\label{prop:anthology-induced-norms}
Let $\Ab\in\RR{m}{n}$, let $p\in[1,\infty]$, and let $p^*$ be its H\"older
conjugate, so that $1/p+1/p^*=1$. Then
\begin{equation}\label{eq:anthology-induced-norms}
  \|\Ab\|_{1\to p}
  = \max_{j\in[n]}\|\Ab_{:,j}\|_p,
  \qquad
  \|\Ab\|_{p\to\infty}
  = \max_{i\in[m]}\|\Ab_{i,:}\|_{p^*}.
\end{equation}
In particular,
\begin{equation}
    \|\Ab\|_{1\to2} =\max_{j\in[n]}\|\Ab_{:,j}\|_2,\quad \|\Ab\|_{2\to\infty}=\max_{i\in[m]}\|\Ab_{i,:}\|_2.\label{eq:1to2_2toinfty-def}
\end{equation}

\end{proposition}

Combining Propositions~\ref{prop:anthology-steepest-descent}
and~\ref{prop:anthology-induced-norms} immediately gives the following identities. 

\begin{corollary}[]
\label{cor:io-normalized-step}
Let $\Gb\in\RR{m}{n}$. The dual norms of $\|\cdot\|_{1\to2}$ and
$\|\cdot\|_{2\to\infty}$ are, respectively,
\begin{align}
  \|\Gb\|_{1\to2}^{\dagger}=\sum_{j=1}^n\|\Gb_{:,j}\|_2,\quad \|\Gb\|_{2\to\infty}^{\dagger}=\sum_{i=1}^m\|\Gb_{i,:}\|_2.
  \label{eq:dual_norms}
\end{align}
\end{corollary}
Define row and column normalization by
\begin{equation}\label{eq:row-normalization-map}
  \bigl(\operatorname{RowNorm}(\Gb)\bigr)_{i,:}
  :=
  \begin{cases}
    \Gb_{i,:}/\|\Gb_{i,:}\|_2, & \Gb_{i,:}\ne\mathbf{0},\\
    \mathbf{0}, & \text{otherwise}.
  \end{cases}
  \qquad
  \operatorname{ColNorm}(\Gb)
  :=\operatorname{RowNorm}(\Gb^\top)^\top.
\end{equation}
Then the $2\to\infty$-penalized problem has the row-normalized solution
\begin{equation}\label{eq:max-row-steepest-descent}
  \Delta\Wb^\star
  =-\lambda\|\Gb\|_{2\to\infty}^{\dagger}
    \operatorname{RowNorm}(\Gb),
\end{equation}
whereas the $1\to2$-penalized problem has the column-normalized solution
\begin{equation}\label{eq:max-column-steepest-descent}
  \Delta\Wb^\star
  =-\lambda\|\Gb\|_{1\to2}^{\dagger}
    \operatorname{ColNorm}(\Gb).
\end{equation}

\subsection{Layer Types and Their Geometries}\label{subsec:layers}

A standard transformer language model contains three qualitatively distinct
layer types, each with a different input/output geometry. We denote $V$ as the vocabulary size and $d$ as the hidden dimension.

\textbf{Hidden linear layers.}
A weight matrix $\Wb \in \RR{n_{\mathrm{out}}}{n_{\mathrm{in}}}$ maps a
\emph{dense} activation $\xb$ into $\yb$, both of which are expected to be
dense vectors. Given a perturbation of the weight $\Delta\Wb$, the bound
$\|\Delta\yb\|_2 \le \twonorm{\Delta\Wb}\|\xb\|_2$ identifies the spectral
norm as the natural penalty~\citep{bernstein2025deriving}, motivating the
Muon polar factor update.

\textbf{Embedding layers.}
\texttt{embed}  ($\Eb \in \RR{d}{V}$) maps token indices to dense
vectors via the lookup $\yb = \Eb\xb$, where $\xb \in \R^V$ is
\emph{one-hot}. Subsection~\ref{sec:embed} reviews the resulting norm from prior work.

\textbf{Language model head.}
\texttt{lm\_head} ($\Lb \in \RR{V}{d}$) maps a hidden state $\yb \in \R^d$ to
logits $\zb = \Lb\yb \in \R^V$ and then to probabilities
$\pb = \softmax(\zb) \in \Delta^V$, where $\Delta^V$ is the probability
  simplex. Given a perturbation $\Delta \Lb$, a good measure of the
  perturbation is not simply $\twonorm{\Delta \zb}$, since its effect must be
  propagated to the probability distribution. Instead, we measure that change by total
  variation, with the analysis in Subsection~\ref{sec:lmhead}.

\section{MuonIO for Vocabulary-Facing Layers}\label{sec:muonio}

The input embedding and output language model head have different storage orientations, but
each has one vector associated with every vocabulary item. We first identify
the appropriate norm for each layer, then orient both matrices along their
vocabulary axis and give a single MuonIO update. Appendix~\ref{app:tightness-proofs} contains the proof of Propositions~\ref{prop:jacobian-tight},~\ref{prop:lipschitz-tight},~\ref{prop:pullback-tight} and Remark~\ref{rem:duality}.

\subsection{Embedding geometry from prior work}\label{sec:embed}

The embedding table $\Eb \in \RR{d}{V}$ acts as a lookup: for a one-hot
token indicator $\xb\in\{0,1\}^V$, the forward pass returns $\Eb\xb$, one
column of $\Eb$. Consequently,
\begin{equation}\label{eq:embed-geometry}
  \max_{\xb \in \{0,1\}^V,\,\sum_i x_i=1}\|\Eb\xb\|_2
  = \max_{t\in[V]}\|\Eb_{:,t}\|_2
  = \|\Eb\|_{1\to 2}.
\end{equation}
The final equality is the $p=2$ case of
Proposition~\ref{prop:anthology-induced-norms}. Thus the $1\to2$ operator norm,
rather than the spectral norm used for dense hidden layers, measures the
largest change to any token embedding.

\citet{bernstein2024old} identify the
$1\to\mathrm{RMS}$ operator norm for embedding layers, which is the same as $1\to2$ operator norm up to a scaling factor of $\sqrt{d}$. We use that established geometry as one half of MuonIO, and our new ingredient is the total variation justification for the matching output layer geometry developed next.

\subsection{Language-model-head geometry}\label{sec:lmhead}

The language model head $\Lb \in \RR{V}{d}$ maps a hidden state $\yb \in \R^d$ to
a probability distribution (as opposed to a dense activation vector) over the vocabulary via
$\pb = \softmax(\Lb\yb) \in \Delta^V$.
We thus identify a matched penalty norm by asking: which norm on $\Delta\Lb$ controls
the worst case change in the \emph{output distribution} induced by the weight
perturbation?

\subsubsection{Total variation stability and the
\texorpdfstring{$2\to\infty$}{2-to-infinity} norm}

A canonical measure of the difference between probability distributions is the total variation distance,
$\|\pb' - \pb\|_{\mathrm{TV}} = \frac{1}{2}\|\pb' - \pb\|_1$.
We show that the $2\to\infty$ operator norm controls this quantity.

\begin{lemma}[Jacobian of softmax]\label{lem:jacobian}
Let $\pb = \softmax(\zb) \in \Delta^V$. The Jacobian of $\softmax$ at $\zb$ is
\begin{equation}
  J(\zb) = \frac{\partial\,\softmax(\zb)}{\partial \zb}
       = \mathrm{diag}(\pb) - \pb\pb^\top.
\end{equation}
\end{lemma}

\begin{proof}
Writing $\pb_i = e^{\zb_i}/\sum_{s=1}^V e^{\zb_s}$, direct differentiation gives
$\partial \pb_i / \partial \zb_j = \pb_i\delta_{ij} - \pb_i \pb_j$.
\end{proof}

\begin{lemma}[$\infty\to 1$ norm of the softmax Jacobian]\label{lem:jacobian-norm}
For every $\zb \in \R^V$,
\begin{equation}\label{eq:jacobian-norm}
  \|J(\zb)\|_{\infty\to 1} \le 1.
\end{equation}
\end{lemma}

\begin{proof}
Fix $\ub \in \R^V$ with $\|\ub\|_\infty \le 1$, and let $\pb = \softmax(\zb)$.
The $i$-th coordinate of $J(\zb)\ub$ is $\pb_i(\ub_i - m)$, where
$m := \pb^\top\ub = \sum_i \pb_i \ub_i$. Hence $ \|J(\zb)\ub\|_1 = \sum_{i=1}^V \pb_i|\ub_i - m|.$
Introduce a random variable $X$ that takes value $\ub_i$ with probability $\pb_i$, so
$m = \ex[X]$ and $\|J(\zb)\ub\|_1 = \ex[|X - \ex[X]|]$.
By Jensen's inequality, $\ex[|X - \ex[X]|] \le \sqrt{\mathrm{Var}(X)}.$ Since $\|\ub\|_\infty \le 1$, all values of $X$ lie in $[-1,1]$.
Popoviciu's inequality on variances~\citep{popoviciu1935equations} states that
$\mathrm{Var}(X) \le (b-a)^2/4$ for any random variable supported on $[a,b]$, so
with $a=-1$ and $b=1$ this gives $\mathrm{Var}(X) \le 1$.
Therefore $\|J(\zb)\ub\|_1 \le 1$, and taking the supremum over all such $\ub$
proves the claim.
\end{proof}

\begin{proposition}[Tightness of Lemma~\ref{lem:jacobian-norm}]\label{prop:jacobian-tight}
    For every $V \ge 2$, inequality~\eqref{eq:jacobian-norm} is attained:
    there exist $\zb \in \R^V$ and $\ub \in \{\pm 1\}^V$ such that
    $\|J(\zb)\ub\|_1 = 1$.
\end{proposition}

\begin{lemma}[Global Lipschitz bound for softmax]\label{lem:lipschitz}
For all $\zb, \zb' \in \R^V$,
\begin{equation}
  \|\softmax(\zb') - \softmax(\zb)\|_1 \le \|\zb' - \zb\|_\infty.
\end{equation}
\end{lemma}

\begin{proof}
Let $\gamma(t) = \zb + t(\zb'-\zb)$ for $t \in [0,1]$. By the fundamental theorem of calculus,
\begin{equation}
  \softmax(\zb') - \softmax(\zb) = \int_0^1 J(\gamma(t))(\zb'-\zb)\,dt.
\end{equation}
Taking norms, applying Lemma~\ref{lem:jacobian-norm} at each $t$, and integrating gives
\begin{equation}
  \|\softmax(\zb') - \softmax(\zb)\|_1
  \le \int_0^1 \|J(\gamma(t))\|_{\infty\to 1}\,\|\zb'-\zb\|_\infty\,dt
  \le \|\zb'-\zb\|_\infty.
\qedhere
\end{equation}
\end{proof}

\begin{proposition}[Tightness of Lemma~\ref{lem:lipschitz}]\label{prop:lipschitz-tight}
The constant $1$ in Lemma~\ref{lem:lipschitz} cannot be improved. Namely, for every
$V \ge 2$,
\begin{equation}\label{eq:lipschitz-tight}
  \sup_{\zb' \ne \zb}\;
  \frac{\|\softmax(\zb') - \softmax(\zb)\|_1}{\|\zb' - \zb\|_\infty} = 1,
\end{equation}
although the supremum is not attained by any finite pair $\zb' \ne \zb$.
\end{proposition}

\begin{lemma}[TV stability of the lm\_head]\label{lem:pullback}
For every hidden state $\yb \in \R^d$ and every perturbation $\Delta\Lb \in \RR{V}{d}$,
\begin{equation}
  \|\softmax((\Lb + \Delta\Lb)\yb) - \softmax(\Lb\yb)\|_{\mathrm{TV}}
  \le \frac{1}{2}\|\Delta\Lb\|_{2\to\infty}\,\|\yb\|_2.
\end{equation}
\end{lemma}

\begin{proof}
Set $\zb = \Lb\yb$ and $\zb' = (\Lb+\Delta\Lb)\yb$, so $\zb'-\zb = (\Delta\Lb)\yb$.
Applying Lemma~\ref{lem:lipschitz} and then the definition of the $2\to\infty$ norm,
\begin{equation}
  \|\softmax(\zb') - \softmax(\zb)\|_1
  \le \|(\Delta\Lb)\yb\|_\infty
  \le \|\Delta\Lb\|_{2\to\infty}\,\|\yb\|_2.
\end{equation}
Dividing by $2$ converts the $\ell_1$ distance to total variation.
\end{proof}

\begin{proposition}[Tightness of Lemma~\ref{lem:pullback}]\label{prop:pullback-tight}
The constant $\frac{1}{2}$ in Lemma~\ref{lem:pullback} cannot be improved: for
every $V \ge 2$ and $d \ge 1$,
\begin{equation}\label{eq:pullback-tight}
  \sup_{\substack{\Lb,\, \Delta\Lb \ne \mathbf{0},\, \yb \ne \mathbf{0}}}\;
  \frac{\|\softmax((\Lb + \Delta\Lb)\yb) - \softmax(\Lb\yb)\|_{\mathrm{TV}}}
       {\frac{1}{2}\|\Delta\Lb\|_{2\to\infty}\,\|\yb\|_2} = 1,
\end{equation}
although the supremum is not attained.
\end{proposition}

Lemma~\ref{lem:pullback} shows that $\|\Delta\Lb\|_{2\to\infty}$ gives sharp
uniform control of worst-case TV perturbation when $\|\yb\|_2$ is held fixed.
The $2\to\infty$ operator norm is therefore the natural penalty for \texttt{lm\_head}.

\subsection{Transposition symmetry}\label{sec:duality}

The two layer geometries are the same after orienting the vocabulary axis
consistently. This follows from the relation between the norms identified in
Sections~\ref{sec:embed} and~\ref{sec:lmhead}.
The observation is a direct consequence of the dual norm
property, and it makes the common $\MuonIO$ update transparent. For completeness, we include the proof in Appendix~\ref{app:tightness-proofs}.

\begin{remark}[Transposition identity]\label{rem:duality}
For any $\Ab \in \RR{m}{n}$ and any $p,q \in [1,\infty]$ with H\"older
conjugates determined by $1/p + 1/p^* = 1$ and $1/q + 1/q^* = 1$,
\begin{equation}\label{eq:duality-general}
  \|\Ab\|_{p\to q} \;=\; \|\Ab^\top\|_{q^*\to p^*}.
\end{equation}
Specializing to $p=2$, $q=\infty$ (so $p^* = 2$ and $q^* = 1$) gives
\begin{equation}\label{eq:duality-special}
  \|\Ab\|_{2\to\infty} \;=\; \|\Ab^\top\|_{1\to 2},
\end{equation}
i.e., the \texttt{lm\_head} norm of $\Lb$ equals the \texttt{embed} norm of $\Lb^\top$.
\end{remark}

\textbf{Consequence under weight tying.}
In language models that tie the embedding and output weights
($\Lb = \Eb^\top$)~\citep{press2017using,inan2016tying},
Remark~\ref{rem:duality} reduces to the tautology
$\|\Eb^\top\|_{2\to\infty} = \|\Eb\|_{1\to 2}$. By
Corollary~\ref{cor:io-normalized-step}, column normalization of $\Eb$ is
exactly row normalization of $\Lb=\Eb^\top$, so $\MuonIO$ acts once on
the shared parameters. The next subsection expresses the same rule for untied
weights.

\subsection{One MuonIO update}\label{subsec:muonio-update}

Put either vocabulary-facing matrix into the canonical row-oriented form
\begin{equation}\label{eq:canonical-io-matrix}
  \Wb =
  \begin{cases}
    \Eb^\top \in \RR{V}{d}, & \text{for the embedding table},\\
    \Lb       \in \RR{V}{d}, & \text{for the language-model head}.
  \end{cases}
\end{equation}
By~\eqref{eq:duality-special}, both layer penalties become the same
$2\to\infty$ norm in this orientation. By
Proposition~\ref{prop:anthology-induced-norms}, this is the maximum 2-norm of a row. Thus both layers use the same local problem
\begin{equation}\label{eq:muonio-obj}
  \Delta\Wb^\star
  \in \arg\min_{\Delta\Wb}\;
  \langle \Gb_k,\Delta\Wb\rangle
  + \frac{1}{2\lambda}\|\Delta\Wb\|_{2\to\infty}^2,
  \qquad \Gb_k=\nabla\mathcal{L}(\Wb_k).
\end{equation}

Corollary~\ref{cor:io-normalized-step} now applies directly, as the unsmoothed
solution row-normalizes $\Gb_k$ and scales it by
$\lambda\|\Gb_k\|_{2\to\infty}^{\dagger}$. In the original storage
orientation, the same rule
normalizes columns of $\Eb$ and rows of $\Lb$. Algorithm~\ref{alg:muonio}
substitutes the momentum estimate for the raw gradient and adds the usual
numerical safeguard  as well as an additional scaling factor $\sqrt{d}$ so that each vector in $\mathbb{R}^d$ has 2-norm $O(\sqrt{d})$.

\begin{algorithm}[ht]
\caption{IO for embedding table or language model head}
\label{alg:muonio}
\textbf{input} row-oriented $\Wb_0\in\RR{V}{d}$ from~\eqref{eq:canonical-io-matrix},
step size $\lambda$, momentum $\beta$, loss $\mathcal{L}$, $\varepsilon>0$ \\
\textbf{initialize} $\Mb_0\gets\mathbf{0}$, $k\gets0$
\begin{algorithmic}[1]
\While{$\Wb_k$ not converged}
  \State Obtain $\tilde{\Gb}_k$ such that
    $\ex[\tilde{\Gb}_k]=\nabla\mathcal{L}(\Wb_k)$
  \State $\Mb_{k+1}\gets\beta\Mb_k+(1-\beta)\tilde{\Gb}_k$
  \State $(\Ub_{k+1})_{j,:}\gets
    \dfrac{(\Mb_{k+1})_{j,:}}
    {\max\{\|(\Mb_{k+1})_{j,:}\|_2,\varepsilon\}}$
    \quad for every $j\in[V]$
  \State $s_{k+1}\gets\sum_{j=1}^V\|(\Mb_{k+1})_{j,:}\|_2$
  \State $\Wb_{k+1}\gets\Wb_k-\lambda s_{k+1}\sqrt{d}\,\Ub_{k+1}$
  \State $k\gets k+1$
\EndWhile
\State \Return $\Wb_k$
\end{algorithmic}
\end{algorithm}

\section{Experiments}\label{sec:experiments}

Our Algorithm~\ref{alg:muonio} acts as a modular component that can be naturally attached to any Muon-style optimizer that treats the I/O layers with AdamW. To demonstrate the effectiveness of our approach, we incorporate IO into Muon~\citep{jordan2024muon} and Polar Express~\citep{amsel2026the}, yielding the corresponding optimizers MuonIO and Polar Express IO. We also include the recent method Ember~\citep{shah2026token} as an additional baseline, as it also provides alternative treatment for the I/O layers.

We evaluate these algorithms along three axes: validation perplexity, optimizer arithmetic cost and optimizer state memory. We additionally report various norms of the gradient for the I/O layers to offer a complementary view of the training dynamics. The central claim is that our geometry-matched update for the I/O layers improves pretraining perplexity while saving roughly $50\%$ of the computation and persistant memory costs compared to AdamW.

\subsection{Experimental Setup}\label{subsec:experiment_setup}

We evaluate all optimizers on autoregressive language model pretraining using the English C4 corpus~\citep{raffel2020exploring}, a cleaned Common Crawl dataset released under the Open Data Commons Attribution License. We train LLaMA-family models from scratch at the 60M, 130M, and 1B scales~\citep{llama2023llama, zhao2024galore}, using untied input embeddings and language model heads. Text is tokenized with the T5-base tokenizer, and individual examples are truncated or padded to 256 tokens. Padding tokens are excluded from the training and evaluation losses. All experiments use bfloat16 precision and an effective batch size of 512 sequences, with gradient accumulation adjusted according to the per-device batch size and GPU count. We use nominal training budgets (including padding positions) of approximately 1.15 billion tokens for the 60M and 130M models and 6.87 billion tokens for the 1B models. Learning rates increase linearly during the first 10\% of training, followed by cosine decay to 10\% of their peak values. Weight decay and gradient clipping are disabled. We evaluate on the held-out C4 validation split using approximately 10 million non-padding tokens per evaluation. Our primary metric is final validation perplexity, computed by exponentiating the average validation loss, with lower values indicating better language-modeling performance. 

We organize the six optimizer configurations into two groups. The first consists of four configurations that use standard Muon for the hidden weight matrices while varying the treatment of the I/O layers. The second examines whether our I/O update extends to other Muon-style optimizers, using Polar Express as a representative example. Across all configurations, auxiliary parameters, including normalization gains and biases when present, are optimized with AdamW. Table~\ref{tab:optimizer-configurations} summarizes the six configurations, where $\eta_a,\eta_h,\eta_i$, and $\eta_o$ denote learning rates. Within each configuration, repeated learning rate symbols indicate that the corresponding parameter groups share a common learning rate during hyperparameter tuning.

\providecommand{\optlr}[2]{%
    \makebox[4.3em][l]{#1}\,$#2$%
}
\providecommand{\optlrwide}[2]{%
    \makebox[6.8em][l]{#1}\,$#2$%
}

\begin{table}[H]
    \caption{Optimizer configurations used in our comparisons.
    Each entry specifies the optimizer and learning rate for the
    corresponding parameter group. Repeated learning rate symbols
    indicate that the same learning rate is shared across parameter
    groups. For each configuration, all distinct learning rates shown
    in the corresponding row are jointly tuned.}
    \label{tab:optimizer-configurations}

    \setlength{\tabcolsep}{5pt}
    \renewcommand{\arraystretch}{1.15}

    \begin{center}
        \begin{tabular}{@{}lllll@{}}
            \toprule
            Configuration
            & Embedding
            & Hidden 2D
            & LM Head
            & Gains \& Biases \\
            \midrule

            Plain Muon (baseline 1)
            & \optlr{AdamW}{\eta_a}
            & \optlrwide{Muon}{\eta_m}
            & \optlr{AdamW}{\eta_a}
            & \optlr{AdamW}{\eta_a} \\

            Muon Tuned (baseline 2)
            & \optlr{AdamW}{\eta_i}
            & \optlrwide{Muon}{\eta_m}
            & \optlr{AdamW}{\eta_o}
            & \optlr{AdamW}{\eta_a} \\

            Ember
            & \optlr{Ember}{\eta_i}
            & \optlrwide{Muon}{\eta_m}
            & \optlr{Ember}{\eta_o}
            & \optlr{AdamW}{\eta_a} \\

            \textbf{MuonIO (ours)}
            & \optlr{\textbf{IO}}{\eta_i}
            & \optlrwide{Muon}{\eta_m}
            & \optlr{\textbf{IO}}{\eta_o}
            & \optlr{AdamW}{\eta_a} \\

            \midrule

            Polar Express Tuned
            & \optlr{AdamW}{\eta_i}
            & \optlrwide{Polar Express}{\eta_h}
            & \optlr{AdamW}{\eta_o}
            & \optlr{AdamW}{\eta_a} \\

            \textbf{Polar Express IO (ours)}
            & \optlr{\textbf{IO}}{\eta_i}
            & \optlrwide{Polar Express}{\eta_h}
            & \optlr{\textbf{IO}}{\eta_o}
            & \optlr{AdamW}{\eta_a} \\

            \bottomrule
        \end{tabular}
    \end{center}
\end{table}

\subsection{Optimizer Performances}
Table~\ref{tab:main-comparison} reports final validation perplexity at all three model scales. 
\begin{table}[H]
    \caption{
        Final validation perplexity on C4 (lower is better).
        The upper block uses standard Muon hidden updates, while
        the lower block uses Polar Express.
        Bold values indicate the lowest mean perplexity within
        each block at each model scale.
        For 60M and 130M, we report the mean and sample standard
        deviation over the three of five seeds with the lowest
        final validation perplexity.
        For 1B, we report the mean and sample standard deviation
        over three seeds.
    }
    \label{tab:main-comparison}

    \begin{center}
        \begin{tabular}{lccc}
            \toprule
            Configuration & 60M & 130M & 1B \\
            \midrule

            Plain Muon (baseline 1)
            & $36.963 \pm 0.054$
            & $28.513 \pm 0.026$
            & $15.116 \pm 0.0003$ \\

            Muon Tuned (baseline 2)
            & $28.502 \pm 0.015$
            & $24.238 \pm 0.008$
            & $14.527 \pm 0.007$ \\

            Ember
            & $28.835 \pm 0.043$
            & $24.074 \pm 0.004$
            & $14.516 \pm 0.032$ \\

            \textbf{MuonIO (ours)}
            & $\mathbf{28.366} \pm 0.025$
            & $\mathbf{23.617} \pm 0.025$
            & $\mathbf{14.374} \pm 0.132$ \\

            \midrule

            Polar Express Tuned
            & $\mathbf{28.385} \pm 0.009$
            & $24.055 \pm 0.010$
            & $14.361 \pm 0.002$ \\

            \textbf{Polar Express IO (ours)}
            & $28.430 \pm 0.025$
            & $\mathbf{23.646} \pm 0.012$
            & $\mathbf{14.160} \pm 0.008$ \\

            \bottomrule
        \end{tabular}
    \end{center}
\end{table}

With standard Muon hidden updates, MuonIO achieves the lowest mean validation perplexity among the four configurations at every model scale, including improvements over the strong baseline Muon Tuned which separately tuned learning rates for all four components of the model. With Polar Express hidden updates, our algorithm lowers mean perplexity at 130M and 1B, while slightly increasing it at 60M. These results show that the improvements at the two larger scales also extend to the Polar Express setting.

\begin{figure}[H]
  \centering
  \includegraphics[width=0.75\linewidth]
    {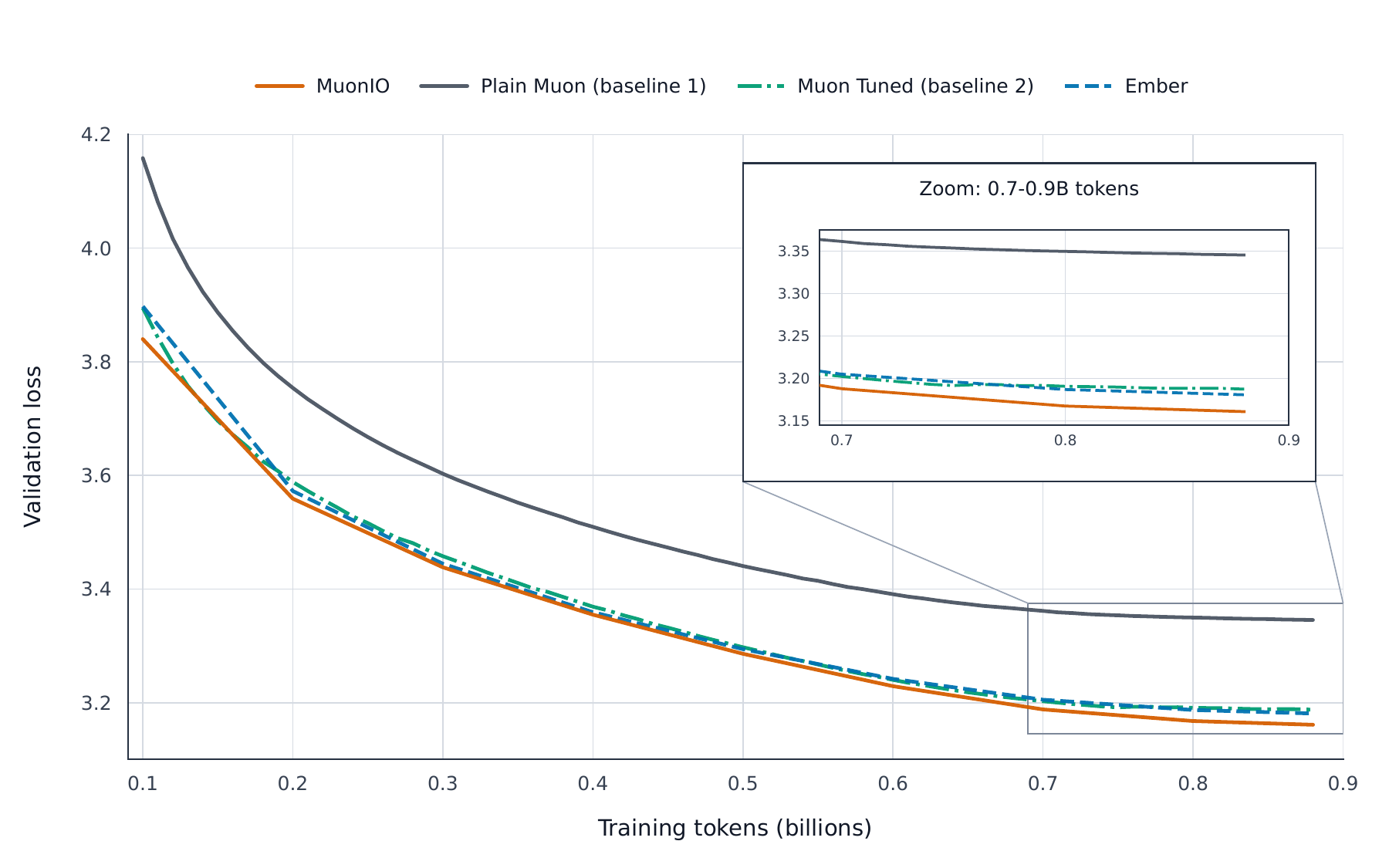}
  \caption{
    Validation loss trajectories on C4 at the 130M scale for Plain Muon (baseline 1), Muon Tuned (baseline 2), Ember and MuonIO. All four configurations use standard Muon hidden updates. The inset highlights the 0.7--0.9B training token range.}
  \label{fig:val-loss}
\end{figure}

Figure~\ref{fig:val-loss} complements the final perplexity
comparison with validation loss trajectories for the four
configurations that used Muon (top 4) for hidden 2D parameters at the 130M scale.

\subsection{IO versus AdamW: exact optimizer cost}
\label{subsec:optimizer-cost}

 Table~\ref{tab:optimizer-cost} compares Algorithm~\ref{alg:muonio}'s exact per-update cost with AdamW on the same $V\times d$ matrix. It needs roughly half the FLOPs and, because it stores only momentum, half the optimizer state. Appendix~\ref{app:complexity} gives the counting convention and the operation-by-operation derivation.

\begin{table}[H]
    \caption{
        FLOPs and persistent optimizer state counts for one
        $V \times d$ I/O matrix and one optimizer update.
        Derivations are in Appendix~\ref{app:complexity}.
    }
    \label{tab:optimizer-cost}

    \begin{center}
        \begin{tabular}{lrr}
            \toprule
            I/O-layer optimizer & FLOPs & Optimizer State \\
            \midrule
            AdamW & $13Vd$ & $2Vd$ entries \\
            IO & $7Vd+3V$ & $Vd$ entries \\
            \bottomrule
        \end{tabular}
    \end{center}
\end{table}

\subsection{Gradient Norm Dynamics}\label{subsec:grad_norm}

\begin{figure}[t!]
    \centering
    \includegraphics[width=0.9\linewidth]
        {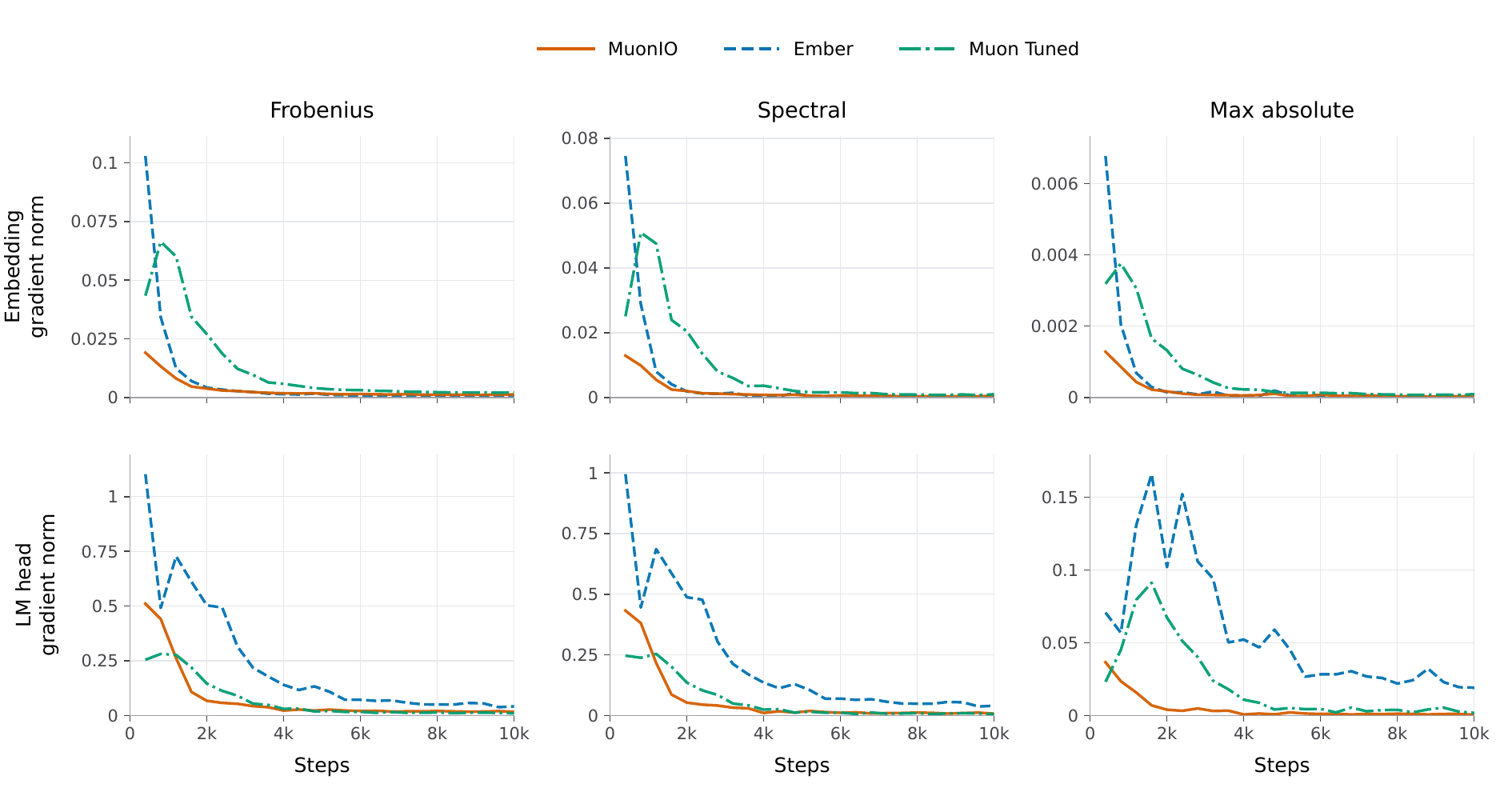}
    \caption{
        \textbf{Gradient magnitudes during 130M model training.}
        Rows show the input embedding table (top) and language
        model head (bottom); columns show the Frobenius norm,
        spectral norm, and maximum absolute gradient entry.
        Curves correspond to MuonIO (orange, solid),
         Ember (blue, dashed), and Muon Tuned (green, dash-dotted). The horizontal axis counts microbatch steps, with eight microbatches per optimizer update.
        Each curve represents one run using the best set of hyperparamters of the corresponding optimizer.
    }
    \label{fig:grad-norm}
\end{figure}

To examine how different optimizers affect the training dynamics of the I/O layers, Figure~\ref{fig:grad-norm} tracks the gradients of these layers during training of the 130M model. We compare three configurations: MuonIO, Muon Tuned, and Ember, all of which use standard Muon updates for the hidden weight matrices and differ only in their treatment of the I/O layers. We consider the gradient both as a matrix and as a flattened vector. From the matrix perspective, we measure its Schatten-\(2\) and Schatten-\(\infty\) norms, corresponding to the Frobenius and spectral norms, respectively. From the vector perspective, we consider the \(\ell_2\) and \(\ell_\infty\) norms. Since the vectorized \(\ell_2\) norm is identical to the Frobenius norm, this yields three distinct quantities: the Frobenius norm, spectral norm, and maximum absolute entry. These norms capture complementary aspects of the gradient magnitude and serve as diagnostics of training progress on the I/O layers.

Across all three metrics, MuonIO drives the I/O gradients down more rapidly than the competing configurations. For the embedding table, MuonIO already exhibits substantially smaller gradient norms at the earliest recorded measurements and quickly approaches near-zero values, whereas Muon Tuned shows an initial increase before decaying and Ember starts with considerably larger gradients. The difference is even more pronounced for the language-model head: MuonIO rapidly reduces all three gradient measures within the first few thousand steps, while Muon Tuned decays more gradually and Ember maintains substantially larger gradients throughout much of the displayed interval. The maximum absolute entry metric shows the clearest separation, with Ember exhibiting large and persistent individual gradient coordinates while MuonIO quickly suppresses them. Overall, these trajectories indicate that, under MuonIO, the embedding and language model head gradients contract substantially earlier in training, particularly for the language-model head.
\section{Conclusion}\label{sec:conclusion}

We have presented $\MuonIO$, one Muon-style update for the two layer types that
the original Muon recipe and many implementations route to AdamW: the input
embedding table and the output language model head. For the embedding table
$\Eb\in\RR{d}{V}$, we adopt the established $1\to\mathrm{RMS}$ geometry
induced by one-hot lookup~\citep{bernstein2024old,bernstein2025modular}, or
equivalently $1\to2$ up to the fixed $d^{-1/2}$ RMS factor. For the output
head $\Lb\in\RR{V}{d}$, our total variation stability analysis of softmax
identifies the $2\to\infty$ geometry. The resulting algorithm $\MuonIO$ is also cheaper than AdamW for vocabulary-facing layers in both optimizer state memory and FLOPs. Our numerical experiments further show that MuonIO improves final loss and perplexity.
\section{Limitations and future work}

Our analysis of $\texttt{lm\_head}$ measures perturbations of the output distribution using total variation and controls them through the $2\rightarrow\infty$ norm. Although this formulation provides a sharp uniform bound and leads to a computationally cheap row-normalization update, it does not explicitly account for the invariance of softmax to constant shifts of all logits. Future work could investigate shift-invariant formulations as well as other measures of distributional change, such as KL divergence, and evaluate whether their optimization benefits justify any additional computational cost.

\bibliography{abc2027_conference}
\bibliographystyle{abc2027_conference}

\appendix
\section{Related Work}
\label{app:related-work}

The original Muon~\citep{jordan2024muon} optimizer uses approximate polar factor updates for hidden
linear layers and assigns embeddings and language model heads to
AdamW. Bernstein and Newhouse~\citep{bernstein2025modular} already derive the embedding geometry used by $\MuonIO$: their $1\to\mathrm{RMS}$ norm formulation, equivalent to $1\to 2$ up to a dimension-dependent factor, gives an update that normalizes each column, though this choice is not empirically evaluated. Scion's~\citep{pethick2025training} operator norm LMO framework includes column normalization
for one-hot inputs and row normalization as an
$\mathrm{RMS}\to\infty$ output option, which is
motivated by control of the largest output coordinate rather than softmax
stability. Separately,
\citet{nair2025softmax} establishes
softmax Lipschitz bounds under matched $\ell_p$ norms. Our analysis instead
uses the mixed $\ell_\infty\to\ell_1$ relation to motivate the
$2\to\infty$ norm for a language model head through total variation
stability.

Several recent optimizers use related normalization directions or otherwise
give special treatment to token-facing matrices. SCALE applies column normalization
to every layer and uses momentum only for the final
layer~\citep{glentis2026scale}. After transposition, its final layer
normalization resembles one component of $\MuonIO$, but SCALE is
otherwise a distinct algorithm motivated by observed gradient variance and
token frequency imbalance. By contrast,
\citet{lau2026symmetry} use a different principle. They require updates to be
equivariant to vocabulary reordering and orthogonal transformations of hidden
space. The resulting family includes the token normalization used here. For
untied heads, their rule also centers momentum, removes common logit shifts,
and chooses scale separately. Ember~\citep{shah2026token} also treats embeddings and language model
heads specially, but follows a different formulation and algorithm using
factorized second moments rather than a Muon-style update.

\section{Proofs}
\label{app:tightness-proofs}

\begin{proof}[\textbf{Proof of Proposition~\ref{prop:jacobian-tight}}]
First suppose $V$ is even. Take $\pb_i = \frac{1}{V}$ for all $i$ (i.e.,
$\zb = \mathbf{0}$). Set $\ub_i = 1$ when $i$ is even and $\ub_i = -1$
when $i$ is odd. Starting from $\|J(\zb)\ub\|_1 = \sum_{i=1}^V \pb_i|\ub_i - m|$, we have $m=0, |\ub_i-m| = 1\ \forall i,$ hence $\|J(\zb)\ub \|_1 = \sum_{i=1}^V \pb_i |\ub_i - m| = \sum_{i=1}^V \frac{1}{V} =1.$

When $V$ is odd, a uniform $\pb$ cannot place mass exactly $\frac{1}{2}$
on the coordinates with $\ub_i = +1$, so we double the mass of one
coordinate instead: take $\zb = (0, \ldots, 0, \ln 2)$, so that
\begin{equation*}
  \pb = \Bigl(\tfrac{1}{V+1}, \ldots, \tfrac{1}{V+1},
  \tfrac{2}{V+1}\Bigr).
\end{equation*}
Choose $\ub \in \{\pm 1\}^V$ with $\ub_V = +1$ and exactly
$\frac{V-3}{2}$ of the first $V-1$ coordinates equal to $+1$ (an integer
since $V$ is odd). The mass on the set $\{i : \ub_i = +1\}$ is then
\begin{equation*}
  \frac{2}{V+1} + \frac{(V-3)/2}{V+1}
  = \frac{4 + V - 3}{2(V+1)}
  = \frac{1}{2},
\end{equation*}
so again $m = 0$ and $|\ub_i - m| = 1$ for all $i$, whence
$\|J(\zb)\ub\|_1 = \sum_{i=1}^V \pb_i = 1$. (Conceptually, this is the
even case with $V+1$ uniform coordinates, two of which --- sharing the
sign $+1$ --- are merged into a single coordinate of double mass.)
\end{proof}

\begin{proof}[\textbf{Proof of Proposition~\ref{prop:lipschitz-tight}}]
\emph{Step 1: the supremum equals $1$.}
Assume first that $V$ is even and let $\ub \in \{\pm 1\}^V$ alternate in sign
as in Proposition~\ref{prop:jacobian-tight}. For $\varepsilon > 0$ set
$\zb = -\frac{\varepsilon}{2}\ub$ and $\zb' = \frac{\varepsilon}{2}\ub$, so that
$\|\zb' - \zb\|_\infty = \varepsilon$. Along the line $\zb(t) = t\ub$, the
softmax takes only two distinct values:
$p_i(t) = \frac{2}{V}\cdot\frac{e^{t}}{e^{t} + e^{-t}}$ on the $V/2$
coordinates with $u_i = +1$, and
$p_i(t) = \frac{2}{V}\cdot\frac{e^{-t}}{e^{t} + e^{-t}}$ on the rest. Summing
the coordinatewise differences between $t = \varepsilon/2$ and
$t = -\varepsilon/2$ gives
\begin{equation}
  \|\softmax(\zb') - \softmax(\zb)\|_1
  = 2\,\frac{e^{\varepsilon/2} - e^{-\varepsilon/2}}
           {e^{\varepsilon/2} + e^{-\varepsilon/2}}
  = 2\tanh(\varepsilon/2),
\end{equation}
so the ratio in~\eqref{eq:lipschitz-tight} equals
$2\tanh(\varepsilon/2)/\varepsilon \to 1$ as $\varepsilon \downarrow 0$.
For $V$ odd, apply the same construction to the first $V-1$ coordinates and set
the last coordinate of both $\zb$ and $\zb'$ to $-M$; letting $M \to \infty$
recovers the same limit. Combined with the upper bound of
Lemma~\ref{lem:lipschitz}, the supremum equals $1$.

\emph{Step 2: the supremum is not attained.}
Suppose equality held in Lemma~\ref{lem:lipschitz} for some finite
$\zb' \ne \zb$. Write $\ub = \zb' - \zb$, let
$\gamma(t) = \zb + t\ub$ for $t \in [0,1]$ be the segment from the proof of
Lemma~\ref{lem:lipschitz}, and let
$\pb(t) := \softmax(\gamma(t)) \in \Delta^V$ denote the softmax along this
segment. Since $t \mapsto \|J(\gamma(t))\ub\|_1$ is continuous and bounded
above by $\|\ub\|_\infty$, equality of the integral bound in the proof of
Lemma~\ref{lem:lipschitz} forces
$\|J(\gamma(t))\ub\|_1 = \|\ub\|_\infty$ for every $t \in [0,1]$. Let $X_t$
take value $u_i$ with probability $p_i(t)$, as in the proof of
Lemma~\ref{lem:jacobian-norm}. Equality in Jensen's inequality requires
$|X_t - \ex[X_t]|$ to be almost surely constant, and equality in Popoviciu's
inequality requires $X_t$ to be supported on
$\{-\|\ub\|_\infty, +\|\ub\|_\infty\}$ with mass $\frac{1}{2}$ each. Together
these give $m(t) := \pb(t)^\top\ub = \ex[X_t] = 0$ and
$\mathrm{Var}(X_t) = \|\ub\|_\infty^2$ for every $t \in [0,1]$. Hence
$m \equiv 0$ and so $m' \equiv 0$. But differentiating along the segment gives
$m'(t) = \ub^\top J(\gamma(t))\ub = \mathrm{Var}(X_t) = \|\ub\|_\infty^2 > 0$, which is a contradiction. Hence the inequality in Lemma~\ref{lem:lipschitz} is
strict for every finite pair, consistent with Step~1, where
$2\tanh(\varepsilon/2) < \varepsilon$ strictly for every $\varepsilon > 0$.
\end{proof}

\begin{proof}[\textbf{Proof of Proposition~\ref{prop:pullback-tight}}]
We show that every logit pair in~\eqref{eq:lipschitz-tight} is realized by an
lm\_head triple with the same ratio. Fix any nonzero $\yb \in \R^d$. For
$\zb, \zb' \in \R^V$ with $\zb' \ne \zb$, define the rank-one matrices
\begin{equation}
  \Lb = \frac{\zb\,\yb^\top}{\|\yb\|_2^2},
  \qquad
  \Delta\Lb = \frac{(\zb' - \zb)\,\yb^\top}{\|\yb\|_2^2},
\end{equation}
so that $\Lb\yb = \zb$ and $(\Lb + \Delta\Lb)\yb = \zb'$. The $t$-th row of
$\Delta\Lb$ is $(\zb' - \zb)_t\,\yb / \|\yb\|_2^2$, whose $\ell_2$ norm is
$|(\zb' - \zb)_t| / \|\yb\|_2$. Proposition~\ref{prop:anthology-induced-norms}
then gives
\begin{equation}
  \|\Delta\Lb\|_{2\to\infty}\,\|\yb\|_2 = \|\zb' - \zb\|_\infty.
\end{equation}
Consequently, the ratio in~\eqref{eq:pullback-tight} evaluated at
$(\Lb, \Delta\Lb, \yb)$ equals the ratio in~\eqref{eq:lipschitz-tight}
evaluated at $(\zb, \zb')$. Taking the supremum over $\zb' \ne \zb$ and
invoking Proposition~\ref{prop:lipschitz-tight}, the supremum
in~\eqref{eq:pullback-tight} is at least $1$; by Lemma~\ref{lem:pullback} it
is at most $1$.

For non-attainment, suppose some triple $(\Lb, \Delta\Lb, \yb)$ with
$\Delta\Lb \ne \mathbf{0}$, $\yb \ne \mathbf{0}$ attained equality
in~\eqref{eq:pullback-tight}. If $(\Delta\Lb)\yb = \mathbf{0}$, the left hand
side of Lemma~\ref{lem:pullback} vanishes while the right hand side is
positive, so equality fails. Otherwise, equality forces in particular
$\|\softmax(\zb') - \softmax(\zb)\|_1 = \|\zb' - \zb\|_\infty$ for the finite
pair $\zb = \Lb\yb \ne \zb' = (\Lb + \Delta\Lb)\yb$, contradicting Step~2 of
the proof of Proposition~\ref{prop:lipschitz-tight}.
\end{proof}

\begin{proof}[\textbf{Proof of Remark~\ref{rem:duality}}]
Using the dual norm characterization
$\|\yb\|_q = \sup_{\|\vb\|_{q^*}\le 1}\langle \vb,\yb\rangle$ and
bilinearity of the inner product,
\begin{align*}
  \|\Ab\|_{p\to q}
  &= \sup_{\|\xb\|_p\le 1} \|\Ab\xb\|_q \\
  &= \sup_{\|\xb\|_p\le 1}\;\sup_{\|\vb\|_{q^*}\le 1}\langle \vb,\Ab\xb\rangle \\
  &= \sup_{\|\vb\|_{q^*}\le 1}\;\sup_{\|\xb\|_p\le 1}\langle \Ab^\top\vb,\xb\rangle \\
  &= \sup_{\|\vb\|_{q^*}\le 1} \|\Ab^\top\vb\|_{p^*} \\
  &= \|\Ab^\top\|_{q^*\to p^*}.
\end{align*}
Specializing to $p=2$, $q=\infty$ (so $p^*=2$, $q^*=1$)
gives~\eqref{eq:duality-special}.
\end{proof}

\section{Exact FLOP and Optimizer-State Accounting}
\label{app:complexity}

This appendix derives the counts reported in
Table~\ref{tab:optimizer-cost}.  Let one input or output parameter matrix have shape $V\times d$ (or its transpose) and let $N := Vd$ be its number of entries. We count FLOPs in the following way.  One scalar addition, subtraction, multiplication, division, comparison or square root counts as one FLOP; consequently, a fused multiply--add counts as two FLOPs. Operations that have in total $O(1)$ FLOPS throughout one optimizer call are ignored.
\paragraph{AdamW.}
Denote the learning rate, weight decay and the betas with $\gamma,\lambda,\beta_1, \beta_2$. Let $\Mb_k$ and $\Vb_k$ denote the first and second moment buffers.  Absorb
the learning rate, bias corrections, and decoupled weight decay into scalars
that are computed once per tensor.  A bias-corrected AdamW update can then be
organized into the four stages in Table~\ref{tab:adamw-flops}.  In the last
two rows, $\Db_k$ denotes the corrected square root denominator.

\begin{table}[H]
    \caption{
        AdamW tensor-arithmetic count for one matrix and one optimizer
        update. The elementwise product and division are denoted by
        $\odot$ and $\oslash$, respectively.
    }
    \label{tab:adamw-flops}

    \begin{center}
        \begin{tabular}{lll}
            \toprule
            Stage & Tensor operation & FLOPs \\
            \midrule

            First moment
            & $\Mb_k \gets
               \frac{\beta_1}{1-\beta_1^t}\Mb_{k-1}
               + \frac{1-\beta_1}{1-\beta_1^t}\Gb_k$
            & $3N$ \\

            Second moment
            & $\Vb_k \gets
               \frac{\beta_2}{1-\beta_2^t}\Vb_{k-1}
               + \frac{1-\beta_2}{1-\beta_2^t}\Gb_k \odot \Gb_k$
            & $4N$ \\

            Corrected denominator
            & $\Db_k \gets \sqrt{\Vb_k} + \varepsilon$
            & $2N$ \\

            Decay and adaptive step
            & $\Wb_{k+1} \gets
               (1-\gamma\lambda)\Wb_k
               - \gamma(\Mb_k \oslash \Db_k)$
            & $4N$ \\

            \midrule
            Total & & $13N = 13Vd$ \\
            \bottomrule
        \end{tabular}
    \end{center}
\end{table}

The first moment update uses two multiplications and one addition per entry. The second moment update additionally squares the gradient, giving three multiplications and one addition per entry.  The corrected denominator uses one square root and one addition per entry.  Finally, the parameter update uses two multiplications, one division, and one subtraction per entry.  Thus the exact FLOP count is $13N$. AdamW keeps both $\Mb_k$ and $\Vb_k$, for $2N$ optimizer state entries.

\paragraph{MuonIO.}
We count one iteration of Algorithm~\ref{alg:muonio} on the row-oriented
matrix $\Wb_k\in\RR{V}{d}$ from~\eqref{eq:canonical-io-matrix}. Write
$r_j:=\|(\Mb_{k+1})_{j,:}\|_2$ for the norm of row $j$. Each $r_j$ is
computed once and reused in both the normalization and the sum
$s_{k+1}$. We then form a single per-row scale
$c_j=a_{k+1}/\max\{r_j,\varepsilon\}$ with $a_{k+1}=\lambda\sqrt{d}\,s_{k+1}$.
This fuses the normalization and scaling into the parameter update,
so $\Ub_{k+1}$ is never materialized.
Table~\ref{tab:muonio-flops} gives the resulting count.

\begin{table}[H]
    \caption{
        MuonIO tensor-arithmetic count for one matrix and one iteration
        of Algorithm~\ref{alg:muonio}. The constant $\lambda\sqrt{d}$
        is computed once, before training.
    }
    \label{tab:muonio-flops}
    
    \begin{center}
        \small
        \begin{tabular}{lll}
            \toprule
            Stage & Operation & FLOPs \\
            \midrule

            Momentum
            & $\Mb_{k+1} \gets
               \beta\Mb_k + (1-\beta)\tilde{\Gb}_k$
            & $3N$ \\

            Row norms
            & $r_j \gets \|(\Mb_{k+1})_{j,:}\|_2$,
              $j \in [V]$
            & $V[d+(d-1)+1] = 2N$ \\

            Sum norms
            & $s_{k+1} \gets \sum_{j=1}^{V} r_j$
            & $V-1$ \\

            Form scalar scale
            & $a_{k+1} \gets (\lambda\sqrt{d})\,s_{k+1}$
            & $1$ \\

            Form row scales
            & $c_j \gets a_{k+1}/\max\{r_j,\varepsilon\}$,
              $j \in [V]$
            & $2V$ \\

            Fused parameter update
            & $(\Wb_{k+1})_{j,:} \gets
               (\Wb_k)_{j,:} - c_j(\Mb_{k+1})_{j,:}$,
              $j \in [V]$
            & $2N$ \\

            \midrule
            Total & & $7N+3V = 7Vd+3V$ \\
            \bottomrule
        \end{tabular}
    \end{center}
\end{table}

For each row of length $d$, the norm takes $d$ squarings, $d-1$ additions, and one square root, or exactly $2d$ FLOPs. Summing over the $V$ rows gives $2Vd=2N$. Forming $c_j$ costs one division and one comparison per row. The fused update, which equals line~6 of Algorithm~\ref{alg:muonio}, then costs one multiplication and one subtraction per matrix entry. The only persistent optimizer tensor is the momentum $\Mb_{k+1}$, so the optimizer state contains $N$ entries.

\section{Experiment Settings and Hyperparameters}\label{app:exp}
We tune optimizer hyperparameters on the C4 validation set, minimizing validation loss or perplexity. For the four-group and IO variants of Polar Express, we use Bayesian optimization with a maximum of 100 trials per sweep and logarithmic learning rate search ranges of \([0.005, 0.08]\) for hidden weights, \([0.001, 0.3]\) for embeddings, \([0.0003, 0.2]\) for the output head, and \([0.0003, 0.03]\) for gains and biases. We search the Polar Express iteration count over \(\{5,6,8\}\). For Muon, we perform a grid search over hidden weight learning rates \(\{0.005,0.01,0.02\}\) and auxiliary learning rates \(\{0.0003,0.0005,0.001\}\), since it has the same learning rates for hidden\_gains\_biases and nonhidden\_params. For Muon Tuned, MuonIO and Ember, we independently search discrete embedding and output head learning rate grids spanning \([0.001,0.3]\) and \([0.0003,0.5]\), respectively, fixing the hidden weight learning rate to \(0.02\) and \(0.04\), respectively. 

\begin{table}[H]
    \caption{
        Optimizer hyperparameters for the 18 selected LLaMA
        pre-training runs on C4. Values use three decimal places;
        nonzero values below $10^{-3}$ use scientific notation.
        $\mu_h$ is hidden weight momentum;
        $(\beta_1^A,\beta_2^A)$ are Adam/AdamW coefficients;
        $\beta_{\mathrm{I/O}}$ is the separate coefficient for non-Adam
        I/O updates. PE denotes Polar Express.
    }
    \label{tab:optimizer-hyperparameters-beta}

    \setlength{\tabcolsep}{3pt}
    \renewcommand{\arraystretch}{1.12}

    \begin{center}
        \small
        \begin{tabular}{@{}llrrrrrrrr@{}}
            \toprule
            Method & Size
            & $\eta_h$
            & $\eta_{i}$
            & $\eta_{o}$
            & $\eta_{a}$
            & $\mu_h$
            & $\beta_1^A$
            & $\beta_2^A$
            & $\beta_{\mathrm{I/O}}$ \\
            \midrule

            Plain Muon
            & 60M  & 0.010 & $3\times 10^{-4}$ & $3\times 10^{-4}$ & 0.001 & 0.950 & 0.900 & 0.950 & -- \\
            & 130M & 0.010 & $3\times 10^{-4}$ & $3\times 10^{-4}$ & 0.001 & 0.950 & 0.900 & 0.950 & -- \\
            & 1B   & 0.019 & $3\times 10^{-4}$ & $3\times 10^{-4}$ & 0.023 & 0.950 & 0.950 & 0.950 & -- \\
            \midrule

            Muon Tuned
            & 60M  & 0.020 & 0.040 & 0.020 & 0.020 & 0.900 & 0.900 & 0.950 & -- \\
            & 130M & 0.020 & 0.040 & 0.020 & 0.020 & 0.900 & 0.900 & 0.950 & -- \\
            & 1B   & 0.016 & 0.197 & 0.004 & 0.015 & 0.900 & 0.900 & 0.950 & -- \\
            \midrule

            Ember
            & 60M  & 0.040 & 0.100 & 0.020 & 0.020 & 0.900 & 0.900 & 0.950 & 0.900 \\
            & 130M & 0.040 & 0.100 & 0.010 & 0.020 & 0.900 & 0.900 & 0.950 & 0.900 \\
            & 1B   & 0.033 & 0.273 & 0.002 & 0.017 & 0.900 & 0.900 & 0.950 & 0.900 \\
            \midrule

            MuonIO
            & 60M  & 0.040 & 0.300 & 0.005 & 0.020 & 0.900 & 0.900 & 0.950 & 0.900 \\
            & 130M & 0.040 & 0.300 & 0.005 & 0.020 & 0.900 & 0.900 & 0.950 & 0.900 \\
            & 1B   & 0.031 & 0.185 & 0.001 & 0.020 & 0.900 & 0.900 & 0.950 & 0.900 \\
            \midrule

            PE Tuned
            & 60M  & 0.030 & 0.047 & 0.004 & 0.012 & 0.900 & 0.900 & 0.950 & -- \\
            & 130M & 0.014 & 0.034 & 0.013 & 0.025 & 0.950 & 0.900 & 0.950 & -- \\
            & 1B   & 0.016 & 0.104 & 0.005 & 0.020 & 0.900 & 0.900 & 0.950 & -- \\
            \midrule

            PE IO
            & 60M  & 0.036 & 0.040 & $9.086\times 10^{-4}$ & 0.026 & 0.900 & 0.900 & 0.950 & 0.900 \\
            & 130M & 0.028 & 0.161 & $3.199\times 10^{-4}$ & 0.014 & 0.900 & 0.900 & 0.950 & 0.900 \\
            & 1B   & 0.026 & 0.056 & 0.002 & 0.017 & 0.900 & 0.900 & 0.950 & 0.900 \\

            \bottomrule
        \end{tabular}

    \end{center}
\end{table}
\paragraph{Parameter groups.}
Table~\ref{tab:optimizer-hyperparameters-beta} lists the selected
optimizer hyperparameters.
The learning rates $\eta_h$, $\eta_{i}$,
$\eta_{o}$, and $\eta_{a}$ correspond to hidden
weights, embeddings, the language model head, and gains/biases,
respectively. Muon Tuned denotes the Muon baseline with separately
tuned I/O learning rates. PE Tuned and PE IO denote the corresponding
Polar Express configurations.

\paragraph{Momentum and Adam coefficients.}
The coefficients $(\beta_1^A,\beta_2^A)$ apply to gains/biases
for every method and additionally to embeddings and the
language model head for Plain Muon, Muon Tuned, and PE Tuned.
For MuonIO and PE IO, $\beta_{\mathrm{I/O}}$ is the I/O momentum
coefficient; their I/O updates use no second moment coefficient.
For Ember, $\beta_{\mathrm{I/O}}$ is the I/O second moment decay
coefficient; its I/O updates use no first moment coefficient.
A dash in the table indicates that the I/O layers use the displayed
Adam coefficient pair. Hidden weight momentum $\mu_h$ is specified
separately.

\paragraph{Shared training settings.}
All runs use C4, a global batch size of 512 sequences, a sequence
length of 256, and bfloat16 precision.
The learning rate follows a cosine decay schedule to 0.1 of its
peak value, with a warmup fraction of 0.1.
The embedding and language model head weights are untied.

\paragraph{Batching.}
We describe each batching configuration by a tuple
(microbatch size per GPU, gradient accumulation steps, number of GPUs),
whose product is the global batch size of 512.
At the 60M scale, runs use $(128,4,1)$, except Muon Tuned and
MuonIO, which use $(64,8,1)$.
At the 130M scale, runs use $(64,8,1)$, except Plain Muon,
which uses $(128,4,1)$.
At the 1B scale, runs use $(64,8,1)$, except MuonIO,
which uses $(64,2,4)$.

\paragraph{Model configuration.}
For the 60M, 130M, and 1B models, respectively, the hidden widths
are $(512,768,2048)$, the numbers of transformer layers are
$(8,12,24)$, the numbers of attention heads are $(8,12,32)$,
and the feed-forward widths are $(1376,2048,5461)$.
All models use an RMSNorm epsilon of $10^{-6}$ and a configured
maximum position count of 2048, while the training sequence
length is 256.

\end{document}